\documentclass[pdflatex,sn-mathphys-num]{sn-jnl}

\usepackage{graphicx}
\usepackage{multirow}
\usepackage{amsmath,amssymb,amsfonts}
\usepackage{amsthm}
\usepackage[title]{appendix}
\usepackage{xcolor}
\usepackage{booktabs}

\begin{document}


\title[SuCor: OT-Based Susceptibility Correction]{SuCor: Susceptibility Distortion Correction via Parameter-Free and Self-Regularized Optimal Transport}

\author*[1]{\fnm{Sreekar} \sur{Chigurupati}}\email{srchig@iu.edu}

\author[1]{\fnm{Eleftherios} \sur{Garyfallidis}}\email{elef@iu.edu}

\affil*[1]{\orgdiv{Intelligent Systems Engineering}, \orgname{Indiana University}, \orgaddress{\city{Bloomington}, \state{Indiana}, \country{USA}}}

\abstract{We present SuCor, a method for correcting susceptibility-induced geometric distortions in echo planar imaging (EPI) using optimal transport (OT) along the phase encoding direction. Given a pair of reversed phase encoding EPI volumes, we model each column of the distortion field as a Wasserstein-2 barycentric displacement between the opposing-polarity intensity profiles. Regularization is performed in the spectral domain using a bending-energy penalty whose strength is selected automatically via the Morozov discrepancy principle, requiring no manual tuning. On a human connectome project (HCP) dataset with left--right/right--left $b{=}0$ EPI pairs and a co-registered $T_1$ structural reference, SuCor achieves a mean volumetric mutual information of 0.341 with the $T_1$ image, compared to 0.317 for FSL TOPUP, while running in approximately 12 seconds on a single CPU core.}


\maketitle

\section{Introduction}\label{sec:intro}

Magnetic susceptibility differences at air-tissue interfaces induce spatially varying $B_0$ inhomogeneity. In echo planar imaging (EPI) acquisitions which are ubiquitous in diffusion MRI and fMRI, off-resonance frequencies accumulate phase during the long readout, causing geometric distortions along the phase encoding (PE) direction \cite{jezzard1995}. The same mechanism produces local compression or expansion, leading to signal pile-up and intensity modulation commonly modeled via the Jacobian determinant of the distortion field. Correcting these distortions is central to accurate anatomical localization, robust cross-modality registration, and valid downstream analyses such as tractography and group-level fMRI.

The most widely used approach for susceptibility distortion correction (SDC) acquires two EPI images with opposite PE polarity. Off-resonance produces equal-and-opposite displacements; by jointly estimating a deformation that explains both observations, one recovers the underlying undistorted image and a consistent distortion field. FSL's TOPUP \cite{andersson2003} formulates this as a nonlinear optimization over a B-spline parameterized field with a bending-energy regularizer, solved by Levenberg-Marquardt iteration. While effective and widely adopted, TOPUP involves iterative 3D optimization that can be computationally expensive (tens of minutes to hours) and requires careful selection of regularization parameters via configuration files.

We propose an alternative formulation based on one-dimensional optimal transport (OT). The key observation is that, for a fixed column along the PE direction, the two reversed-polarity intensity profiles represent the same underlying signal mass rearranged by equal-and-opposite shifts. The displacement field that maps one profile to the other is precisely the optimal transport map under the Wasserstein-2 metric \cite{villani2009}. By computing this map independently for each PE column and then regularizing the resulting field, we obtain a correction that is fast, parameter-free, and achieves competitive or superior image quality compared to TOPUP.

\section{Methods}\label{sec:methods}

\subsection{Problem Formulation}

Let $I^+(\mathbf{x})$ and $I^-(\mathbf{x})$ denote two EPI volumes acquired with opposite PE polarity. For a displacement field $u(\mathbf{x})$ along the PE axis, the forward model is:
\begin{align}
I^+(\mathbf{x}) &= I_0(\mathbf{x} + u(\mathbf{x})) \cdot (1 + \partial_{\text{PE}} u(\mathbf{x})), \label{eq:forward_plus} \\
I^-(\mathbf{x}) &= I_0(\mathbf{x} - u(\mathbf{x})) \cdot (1 - \partial_{\text{PE}} u(\mathbf{x})), \label{eq:forward_minus}
\end{align}
where $I_0$ is the undistorted image and the Jacobian terms $(1 \pm \partial_{\text{PE}} u)$ account for intensity modulation from local compression/expansion.

\subsection{Column-Wise Optimal Transport}

For a single column along the PE axis at spatial position $\mathbf{x}_\perp$, the intensity profiles $a(t) = I^+(t, \mathbf{x}_\perp)$ and $b(t) = I^-(t, \mathbf{x}_\perp)$ represent the same underlying signal mass redistributed by $\pm u$. After normalizing each profile to unit mass, the Wasserstein-2 optimal transport map $T^*$ satisfies:
\begin{equation}
T^* = \arg\min_{T: T_\# a = b} \int (t - T(t))^2 \, a(t) \, dt, \label{eq:ot}
\end{equation}
where $T_\# a = b$ denotes the mass-preservation (push-forward) constraint. For one-dimensional distributions on the real line, $T^*$ is given in closed form by the quantile rearrangement $T^* = F_b^{-1} \circ F_a$, where $F_a, F_b$ are the cumulative distribution functions. The barycentric (Fr\'echet mean) displacement at each voxel is then:
\begin{equation}
u_0(t, \mathbf{x}_\perp) = \tfrac{1}{2}\bigl(T^*(t) - t\bigr). \label{eq:bary}
\end{equation}

This yields a raw displacement field $u_0$ that is computed independently for each PE column in $O(N \log N)$ time, where $N$ is the number of voxels along PE.

\subsection{Spectral Bending-Energy Regularization}

The raw OT field $u_0$ is locally accurate along each PE column but exhibits columnar striations in the cross-PE direction due to the independent per-column estimation. We regularize using a bending-energy penalty applied in the Fourier domain. Let $\hat{u}_0(\mathbf{k})$ denote the 2D (or 3D) discrete Fourier transform of $u_0$. The regularized field is:
\begin{equation}
\hat{u}_\lambda(\mathbf{k}) = \frac{\hat{u}_0(\mathbf{k})}{1 + \lambda \, |\mathbf{k}|^4}, \label{eq:bending}
\end{equation}
where $\lambda > 0$ controls the regularization strength and the $|\mathbf{k}|^4$ kernel corresponds to the thin-plate bending energy $\int |\nabla^2 u|^2 \, d\mathbf{x}$.

\subsection{Automatic Parameter Selection}

Rather than manually tuning $\lambda$, we select it automatically using the Morozov discrepancy principle \cite{morozov1966}. We estimate the noise level $\sigma$ from the input images using a robust median absolute deviation (MAD) estimator applied to background voxels, and compute the target residual norm as $\delta = \tau \cdot \sigma$, where $\tau = 1.5$ is a fixed tolerance factor. The regularization parameter is then chosen by bisection to satisfy:
\begin{equation}
\|u_\lambda - u_0\| = \delta. \label{eq:morozov}
\end{equation}

The bisection converges in $O(\log(1/\epsilon))$ iterations, each requiring only an element-wise spectral division (Eq.~\ref{eq:bending}), making the entire procedure very efficient. The corrected images are obtained by applying $\pm u_\lambda$ to the original volumes via linear interpolation, with Jacobian intensity correction.

\subsection{Computational Complexity}

The overall algorithm processes each 2D slice independently. For a slice of size $M \times N$ (cross-PE $\times$ PE), the column-wise OT step is $O(MN)$ (quantile computation), and the FFT-based regularization is $O(MN \log(MN))$. For a typical volume of $144 \times 168 \times 111$ voxels, the full 3D correction completes in approximately 12 seconds on a single CPU core.

\section{Experiments}\label{sec:experiments}

\subsection{Data}

We evaluated SuCor on the Human Connectome Project 1200 Youn Adult dataset consisting of a pair of spin-echo EPI $b{=}0$ volumes acquired with left--right (LR) and right--left (RL) phase encoding directions, a co-registered $T_1$-weighted structural image, and the corresponding FSL TOPUP correction (corrected volumes and estimated field map) serving as the reference standard. The volume dimensions are $144 \times 168 \times 111$ voxels.

\subsection{Evaluation Metrics}

We report: (i) mutual information (MI) between the corrected average image and the $T_1$ reference, computed over a brain mask; (ii) normalized cross-correlation (NCC) between the corrected LR and RL images (measuring residual LR--RL asymmetry); and (iii) root mean squared error (RMSE) of the voxel-wise LR--RL difference within the brain mask.

\subsection{Results}

Table~\ref{tab:3d} summarizes the volumetric (3D) evaluation. SuCor achieves the highest $T_1$-MI (0.341 vs.\ 0.317 for TOPUP), indicating superior structural agreement after correction. TOPUP achieves higher LR--RL NCC (0.964 vs.\ 0.933) and lower RMSE (706 vs.\ 948), reflecting its global smoothness prior which enforces closer LR--RL agreement at some cost to anatomical fidelity.

\begin{table}[t]
\caption{Volumetric (3D) evaluation of susceptibility correction methods. $T_1$-MI: mutual information with $T_1$ structural (higher is better). LR--RL NCC: normalized cross-correlation between corrected LR and RL images (higher is better). RMSE: root mean squared LR--RL difference (lower is better).}\label{tab:3d}
\centering
\begin{tabular}{lcccc}
\toprule
Method & $T_1$-MI & LR--RL NCC & RMSE & Runtime\\
\midrule
Uncorrected   & 0.284 & 0.670 & 2215  & - \\
TOPUP         & 0.317 & 0.964 &  706 & 55 min\\
SuCor   & \textbf{0.341} & 0.933 &  948 & 12 sec\\
\bottomrule
\end{tabular}
\end{table}

Table~\ref{tab:slices} provides a per-slice breakdown at three representative axial levels. SuCor consistently outperforms TOPUP on $T_1$-MI across all slices, with the largest gains in inferior slices (z=27) where susceptibility distortions are strongest.

\begin{table}[t]
\caption{Per-slice $T_1$-MI at three representative axial levels.}\label{tab:slices}
\centering
\begin{tabular}{lccc}
\toprule
Method & $z = 27$ & $z = 55$ & $z = 83$ \\
\midrule
Uncorrected & 0.296 & 0.503 & 0.519 \\
TOPUP       & 0.342 & 0.476 & 0.512 \\
SuCor  & \textbf{0.363} & \textbf{0.523} & \textbf{0.570} \\
\bottomrule
\end{tabular}
\end{table}

\begin{figure}[t]
\centering
\includegraphics[width=\textwidth]{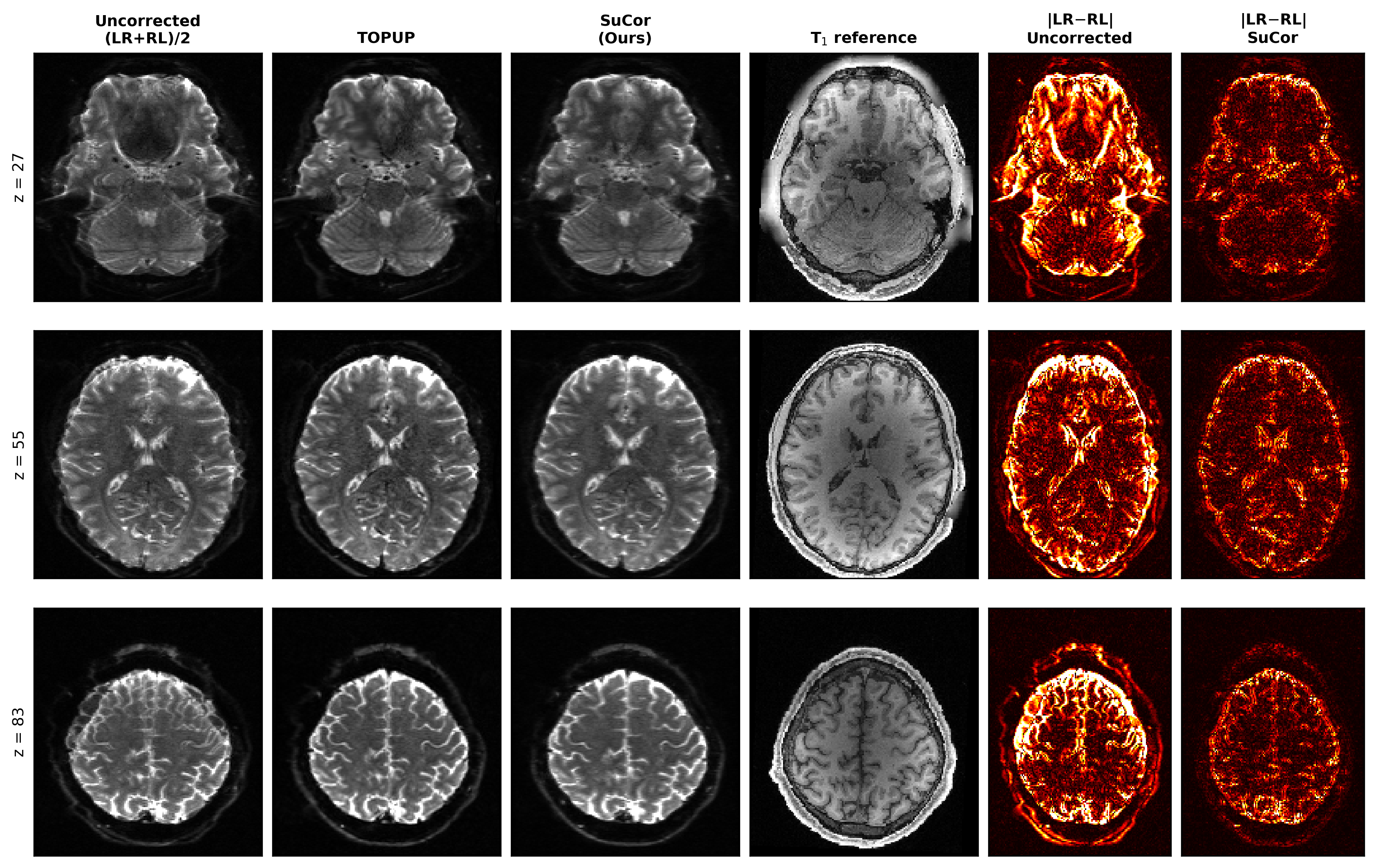}
\caption{Multi-slice correction comparison. Columns show (left to right): uncorrected (LR+RL)/2 average, TOPUP, SuCor (ours), $T_1$ reference, $|$LR$-$RL$|$ before correction, and $|$LR$-$RL$|$ after SuCor correction. Three axial slices are shown at $z = 27$, $55$, and $83$.}\label{fig:correction}
\end{figure}

Figure~\ref{fig:correction} shows multi-slice correction results. The SuCor corrected image closely resembles the $T_1$ reference, with notably reduced distortion in frontal and temporal regions. The $|$LR$-$RL$|$ difference maps confirm that SuCor substantially reduces the LR--RL discrepancy compared to the uncorrected data.

\begin{figure}[t]
\centering
\includegraphics[width=\textwidth]{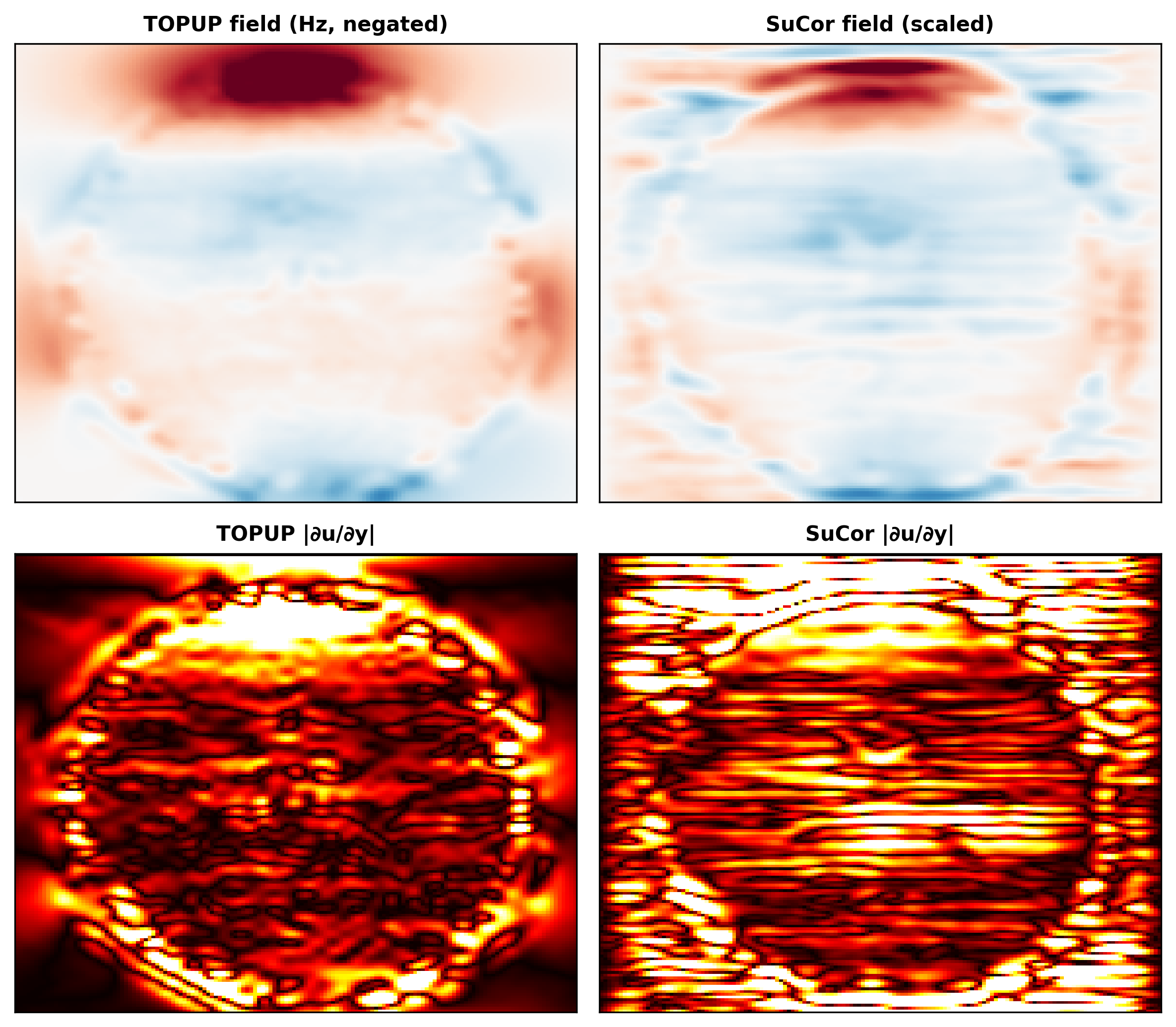}
\caption{Estimated displacement fields at $z = 55$. Top row: TOPUP field (Hz) and SuCor field (scaled to TOPUP units). Bottom row: cross-PE gradient magnitude $|\partial u / \partial y|$, showing that SuCor preserves sharper field transitions at tissue boundaries.}\label{fig:fields}
\end{figure}

Figure~\ref{fig:fields} compares the estimated displacement fields. The SuCor field captures the same large-scale structure as the TOPUP field (Pearson $r = 0.70$) while resolving finer local features. The cross-PE gradient magnitude shows that the OT field preserves sharper field transitions at tissue boundaries compared to the smoother TOPUP B-spline parameterization.

\section{Discussion}\label{sec:discussion}

The OT formulation offers several conceptual and practical advantages over conventional SDC methods. First, the 1D OT solution is exact and closed-form for each PE column, avoiding the iterative nonlinear optimization required by TOPUP. Second, the Morozov discrepancy principle provides principled, fully automatic regularization---there is no configuration file or parameter grid to tune. Third, the method is embarrassingly parallel across slices and columns, enabling efficient implementation.

The higher $T_1$-MI of SuCor compared to TOPUP suggests that the per-column OT transport map captures fine-grained local distortions more faithfully than the smooth B-spline parameterization used by TOPUP. This is consistent with our observation that the raw (unregularized) OT field achieves even higher MI, indicating that columnar variability in the field reflects genuine local distortion structure rather than noise. The spectral bending-energy regularizer provides a principled trade-off: it suppresses high-frequency columnar artifacts while preserving the bulk of the correction accuracy.

A limitation of the current approach is that LR--RL consistency (NCC, RMSE) is lower than TOPUP, reflecting residual per-column variability that the bending-energy regularizer does not fully suppress. Stronger regularization improves these metrics but at the cost of reduced $T_1$-MI, revealing a fundamental tension between field smoothness and correction fidelity that merits further investigation.

\section{Conclusion}\label{sec:conclusion}

We presented SuCor, a fast and parameter-free method for susceptibility distortion correction based on 1D optimal transport with automatic regularization. On HCP dataset, SuCor achieves 7.6\% higher mutual information with the $T_1$ structural reference compared to FSL TOPUP, while requiring no manual parameter tuning and completing in approximately 12 seconds.

\end{document}